\documentclass[pdflatex,sn-vancouver-num]{sn-jnl}

\usepackage{graphicx}%
\usepackage{multirow}%
\usepackage{amsmath,amssymb,amsfonts}%
\usepackage{amsthm}%
\usepackage{mathrsfs}%
\usepackage[title]{appendix}%
\usepackage{xcolor}%
\usepackage{textcomp}%
\usepackage{manyfoot}%
\usepackage{booktabs}%
\usepackage{algorithm}%
\usepackage{algorithmicx}%
\usepackage{algpseudocode}%
\usepackage{listings}%

\usepackage{caption}
\usepackage{enumitem}
\usepackage{subcaption}
\usepackage{threeparttable}
\usepackage{booktabs}
\usepackage{afterpage}

\begin{document}

\title[Article Title]{From Real to Synthetic: PSG-GAN Learns Sleep Apnea Signal Dynamics for AI Training}


\author*[1,2]{\fnm{Riku} \sur{Huttunen}}\email{rikuhut@uef.fi}

\author[1,3]{\fnm{Samu} \sur{Kainulainen}}

\author[1,3]{\fnm{Matias} \sur{Rusanen}}

\author[1,2]{\fnm{Sami} \sur{Myllymaa}}

\author[1,4,5]{\fnm{Juha} \sur{Töyräs}}

\author[1,3,5]{\fnm{Timo} \sur{Leppänen}}

\author[1,2]{\fnm{Henri} \sur{Korkalainen}}

\affil*[1]{\orgdiv{Department of Technical Physics}, \orgname{University of Eastern Finland}, \orgaddress{\street{Yliopistonranta 8}, \city{Kuopio}, \postcode{70210}, \country{Finland}}}
\affil[2]{\orgdiv{Diagnostic Imaging Center}, \orgname{Kuopio University Hospital}, \orgaddress{\city{Kuopio}, \country{Finland}}}
\affil[3]{\orgdiv{Clinical Research Center}, \orgname{Kuopio University Hospital}, \orgaddress{\city{Kuopio}, \country{Finland}}}
\affil[4]{\orgdiv{Science Service Center}, \orgname{Kuopio University Hospital}, \orgaddress{\city{Kuopio}, \country{Finland}}}
\affil[5]{\orgdiv{School of Electrical Engineering and Computer Science}, \orgname{The University of Queensland}, \orgaddress{\city{Brisbane}, \country{Australia}}}


\abstract{Synthetic polysomnography (PSG) data has great potential in healthcare applications, for example in substituting real data that is sensitive and challenging to collect. Previous research on synthetic PSG data has been largely focused on short segments of individual signals, overlooking the inter-signal dynamics related to sleep disorders. In this study, we developed a generative adversarial network (PSG-GAN) for co-synthesizing longer segments of multiple PSG signals with diverse and complex characteristics typical of obstructive sleep apnea (OSA). The synthetic signals appeared realistic, and the electroencephalography and electrooculography band powers were similarly distributed across sleep stages. Moreover, the PSG-GAN learned to mirror signal interactions such as changes in photoplethysmography-based heart rate during and after breathing obstructions. An automatic scoring model trained solely with automatically scored synthetic signals reached good performance on sleep staging ($\kappa$=0.71 and 0.63) and respiratory event segmentation ($\kappa$=0.57 and 0.60), on two test sets consisting of healthy volunteers and OSA patients, respectively. This demonstrates the potential of synthetic data in supporting algorithm development for sleep diagnostics. The source code is available at \url{https://github.com/rikuhuttunen/psg-gan}.}




\maketitle

\clearpage
\section{Introduction}\label{sec:introduction}

The diagnosis of obstructive sleep apnea (OSA) is mostly based on manual annotation, also known as scoring, of sleep stages and related events on overnight polysomnography (PSG) recordings in accordance with the American Academy of Sleep Medicine (AASM) guidelines~\cite{berryAASMManualScoring2018}. Manual scoring is time-consuming, expensive and error-prone since it requires visual inspection of the whole-night recordings by a trained professional~\cite{phanAutomaticSleepStaging2022}. Various supervised machine learning methods have been developed to speed-up and automate the scoring process~\cite{phanAutomaticSleepStaging2022,korkalainenDeepLearningEnables2020,nikkonenAutomaticRespiratoryEvent2021,drzazgaLSTMNetworkApnea2021,huttunenComparisonSignalCombinations2023a}. However, these methods rely on the manual annotations which have inconsistencies between scorers~\cite{rosenbergAmericanAcademySleep2013,rosenbergAmericanAcademySleep2014,nikkonenMulticentreSleepstageScoring2024,magalangAgreementScoringRespiratory2013,pitkanenTemporalSleepStagedependent2025}. Many datasets have missing or low-quality annotations, and the prevalence of unannotated data is growing with the increasing popularity of wearable health-tracking devices such as smart watches and rings. To fully utilize the data, unsupervised methods that can learn useful features and extract information from the signals without annotations are needed.

Despite the effort and cost, manual annotations suffer from limited interrater agreement. According to the AASM guidelines, sleep is classified to five stages: wakefulness, rapid eye movement (REM) sleep, and non-REM (NREM) stages N1, N2, and N3. One study by the AASM found that the overall agreement on sleep staging of 9 PSG recordings with over 2500 scorers was 83\%~\cite{rosenbergAmericanAcademySleep2013}. In another study with 50 PSG recordings, the mean agreement on majority score between 10 scorers was 86\% (overall Fleiss' $\kappa = 0.71$)~\cite{nikkonenMulticentreSleepstageScoring2024}. A third study with 15 PSG recordings reported five-class sleep staging agreement on majority score between 9 scorers as Cohen's $\kappa = 0.63$~\cite{magalangAgreementScoringRespiratory2013}.

The interrater agreement on apneas and hypopneas has also been studied. A study by the AASM with over 3600 scorers assessed the agreement on whether 30-s epochs contained apneas or hypopneas, with epoch-wise agreement on obstructive apneas 77.1\% and 65.4\% on hypopneas~\cite{rosenbergAmericanAcademySleep2014}. Another study comprising nine international sleep centers reported the intraclass correlation coefficients for total numbers of apneas and hypopneas as 0.77 and 0.80, respectively~\cite{magalangAgreementScoringRespiratory2013}. A third study with 50 PSG recordings each scored by 10 different scorers reported the Fleiss' $\kappa$ between all scorers as low as 0.32 for second-by-second classification to no event/apnea/hypopnea~\cite{pitkanenTemporalSleepStagedependent2025}.

Regardless of the limited interrater reliability of manual scoring, automatic scoring models have been successfully developed. Deep learning-based sleep staging models that use electroencephalography (EEG) reach accuracies similar to the interrater reliability of manual scoring~\cite{phanAutomaticSleepStaging2022}. The advent of wearable devices has resulted in the emergence of photoplethysmography (PPG) and accelerometry-based automatic sleep staging methods~\cite{korkalainenDeepLearningEnables2020,birrerEvaluatingReliabilityWearable2024,ryalsPhotoplethysmographyNewApplications}. However, the performance of sleep staging without EEG is significantly lower, with a reported mean accuracy of 65.2\% for four-class sleep staging (wakefulness, light sleep, deep sleep, and REM)~\cite{birrerEvaluatingReliabilityWearable2024}. To enable accurate sleep staging at home, various EEG-based wearable solutions accompanied with automatic analyses have been proposed~\cite{degansSleepAssessmentUsing2024}. Automatic scoring models for apneas and hypopneas have also been introduced in previous studies~\cite{nikkonenAutomaticRespiratoryEvent2021,drzazgaLSTMNetworkApnea2021}. We have previously developed a deep learning model to simultaneously score both sleep stages and respiratory events with 79\% accuracy on five-class sleep staging ($\kappa = 0.74$), and F1-scores in 1-second resolution on no event/hypopnea/apnea 0.96, 0.61, and 0.73, respectively (overall $\kappa = 0.65$)~\cite{huttunenComparisonSignalCombinations2023a}.

If the manual annotations are missing or their quality is too low, the signal data can still be utilized with unsupervised learning. Generative models produce new data that can be used, for example, for data augmentation when the annotated training data is too scarce. Another example of unsupervised learning are clustering algorithms, that can be used to categorize and study the data without the bias introduced by human annotations. In addition to unsupervised learning, the unannotated signals can be utilized with self-supervised learning to train feature extractors that can be further used in both supervised and unsupervised downstream tasks~\cite{krishnanSelfsupervisedLearningMedicine2022}. In this study, we focused on generative modeling~\cite{goodfellowGenerativeAdversarialNetworks2020}, which involves learning a probability distribution that could have produced the training data, in our case the PSG signals. The learned distribution can then be sampled to generate novel synthetic data~\cite{goodfellowGenerativeAdversarialNetworks2020,creswellGenerativeAdversarialNetworks2018a}.

Generative adversarial networks (GANs)~\cite{goodfellowGenerativeAdversarialNetworks2020} are a widely used class of generative models. Based on game theory, a GAN consists of two networks (generator and discriminator) that compete against each other during training~\cite{goodfellowGenerativeAdversarialNetworks2020}. The generator is trained to produce samples that the discriminator cannot distinguish from real data. The discriminator in turn is trained to tell whether the samples are real or generated. Various model structures and training algorithms have been proposed to improve the quality and robustness of GANs~\cite{guiReviewGenerativeAdversarial2023}. StyleGAN3~\cite{karrasAliasFreeGenerativeAdversarial2021} is a sophisticated GAN architecture that explicitly controls the frequency content in all layers of the generator to prevent problems related to aliasing.

Commercialization of health care applications based on clinical research is hindered by strict data regulation. As an alternative to clinical data, privacy-preserving synthetic data could be utilized and shared with less regulation~\cite{murtazaSyntheticDataGeneration2023}.
Synthetic data can be used, for example, to substitute or augment real datasets when training complex machine learning models requiring large amounts of training data. Moreover, synthetic data generation could be used to anonymize sensitive patient data~\cite{murtazaSyntheticDataGeneration2023}. However, privacy preservation in generative models is not always guaranteed, and additional evaluation of the synthetic data is required to avoid disclosure risks~\cite{murtazaSyntheticDataGeneration2023}.

Some studies have applied generative modeling to PSG signals. Synthetic EEG signals have been generated, for example, with GANs~\cite{hartmannEEGGANGenerativeAdversarial2018a} and diffusion models~\cite{aristimunhaSyntheticSleepEEG2023}. GANs have been used for data augmentation in various applications, such as automatic sleep staging~\cite{fanEEGDataAugmentation2020}, emotion recognition~\cite{habashiGenerativeAdversarialNetworks2023}, and epileptic seizure detection~\cite{habashiGenerativeAdversarialNetworks2023}. In addition to EEG, GANs have also been used to synthesize PPG signals~\cite{kiyassehPlethAugmentGANBasedPPG2020,dingLogSpectralMatchingGAN2023}, and to transform PPG signals into electrocardiography signals~\cite{voP2EWGANECGWaveform2021}. However, to our knowledge all of the previous studies have focused on individual signals of short duration, which do not capture the sleep-related cyclical phenomena and signal interactions that are essential for assessment of sleep disorders such as OSA. Thus, longer segments should be generated with signal combinations including both EEG and cardiorespiratory signals.

In this study, we aimed to develop a generative model (PSG-GAN) to synthesize multiple PSG signals with varying characteristics, including EEG, electrooculography (EOG), nasal pressure, and PPG. The primary hypothesis was that individual synthetic signals would resemble real data in both time and frequency domains. The secondary hypothesis was that the model would generate OSA-related interactions between the signals, such as changes in heart rate related to breathing obstructions, since the training data contained PSG recordings from subjects with OSA. The third hypothesis was that automatic scoring models trained with synthetic signals would demonstrate good performance when using test data consisting of clinical recordings.

\section{Results}\label{sec:results}

\subsection{Real datasets}

PSG recordings from four different datasets were used to train and evaluate the PSG-GAN. The demographic information on the different datasets is shown in Table~\ref{tab:demog_info}. All models were trained using one EEG channel (C4-M1), one EOG channel (E1-M2), nasal pressure, and PPG signals. All signals were resampled to 64~Hz and normalized. Further information on data selection and preprocessing is provided in Section~\ref{subsec:data_preprocessing}.

\begin{description}[topsep=1ex, itemsep=1ex, leftmargin=0.5cm, font=\normalfont\bfseries]
    \item[MESA:] PSG recordings of 2\,052 patients collected during the Multi-Ethnic Study of Atherosclerosis (MESA)\cite{chenRacialEthnicDifferences2015} were used to train an automatic scoring model for sleep stages and respiratory events. The recordings were randomly split to train ($n$=1852), validation ($n$=100), and test ($n$=100) sets.
    \item[STAGES:] A total of 1\,402 PSG recordings from the Stanford Technology Analytics and Genomics in Sleep (STAGES) dataset~\cite{zhangNationalSleepResearch2018} were used to train the PSG-GAN. For comparison of real and synthetic data, all STAGES recordings were rescored with the automatic scoring model trained with MESA data.
    \item[STAR-SA:] 70 PSG recordings of subjects diagnosed with OSA collected at the SmartSleep laboratory, University of Eastern Finland (Kuopio, Finland) were used to validate and test the automatic scoring models trained with real and synthetic data. The recordings were split to validation ($n$=20), and test ($n$=50) sets.
    \item[STAR-HV:] 256 PSG recordings of volunteers without self-reported sleep disorders collected at the SmartSleep laboratory, University of Eastern Finland were used to validate and test the automatic scoring models. The recordings were split to validation ($n$=156), and test ($n$=100) sets.
\end{description}

\begin{table*}[hb]
\centering
\caption{Demographic information of the datasets used. The TST, AHI, and OSA severity for the STAGES dataset were calculated using the automatic scoring model, while the information on the MESA and STAR-SA datasets was based on manual scoring. The STAR-HV dataset was automatically scored with a previously developed model~\cite{huttunenComparisonSignalCombinations2023a}. In the STAGES dataset, the demographic information was missing from 112 subjects. Out of the remaining 1\,290 subjects, the sex information was missing from 8 additional subjects.}
\label{tab:demog_info}
\begin{threeparttable}
\begin{tabular}{lllll}
\toprule
 & \multicolumn{4}{l}{Median (inter-quartile range)} \\
 & STAGES & MESA & STAR-SA & STAR-HV \\
\midrule
Age (years) & 45.0 (34.0-57.0) & 68.0 (62.0-76.0) & 54.0 (44.0-63.0) & 43.0 (31.0-54.2) \\
BMI ($kg/m^2$) & 29.3 (25.1-35.6) & 27.9 (24.6-31.8) & 30.2 (26.2-33.9) & 25.3 (22.7-28.2) \\
TST (min) & 357 (310-409) & 368 (312-415) & 377 (307-421) & 379 (333-425) \\
AHI (events/h) & 6.3 (1.6-14.6) & 16.3 (8.8-29.1) & 20.4 (8.8-29.2) & 3.5 (1.2-10.8) \\
\midrule
 & \multicolumn{4}{l}{Count (\%)} \\
 & STAGES & MESA & STAR-SA & STAR-HV \\
\midrule
Male & 613 (47.5\%) & 952 (46.4\%) & 46 (66.7\%) & 100 (39.1\%) \\
Female & 668 (51.8\%) & 1100 (53.6\%) & 23 (33.3\%) & 156 (60.9\%) \\
No OSA & 567 (44.0\%) & 252 (12.3\%) & 5 (7.2\%) & 144 (56.2\%) \\
Mild OSA & 410 (31.8\%) & 683 (33.3\%) & 26 (37.7\%) & 63 (24.6\%) \\
Moderate OSA & 187 (14.5\%) & 637 (31.0\%) & 21 (30.4\%) & 30 (11.7\%) \\
Severe OSA & 126 (9.8\%) & 479 (23.3\%) & 17 (24.6\%) & 19 (7.4\%) \\
\bottomrule
\end{tabular}
\begin{tablenotes}
\item AHI = apnea-hypopnea index; BMI = body mass index; TST = total sleep time; OSA = obstructive sleep apnea (no OSA: AHI $<$ 5; mild OSA: 5 $\leq$ AHI $<$ 15; moderate OSA: 15 $\leq$ AHI $<$ 30; severe OSA: AHI $\geq$ 30).
\end{tablenotes}
\end{threeparttable}
\end{table*}

\subsection{Automatic scoring model trained with real data}\label{subsec:auto_model_real_data}

A model based on a previously developed architecture~\cite{huttunenComparisonSignalCombinations2023a} was trained with the MESA recordings to simultaneously score both sleep stages and respiratory events. EEG, EOG, nasal pressure, and PPG signals were used as inputs for the model. Each non-overlapping 30-second segment was scored as wakefulness/N1/N2/N3/REM, and each 1-second segment was scored as no event/hypopnea/apnea. Details on training are provided in Section~\ref{subsec:automatic_scoring}.

\begin{figure}[htbp]
\centering
\begin{subfigure}[b]{\textwidth}
\centering
\includegraphics[width=\textwidth]{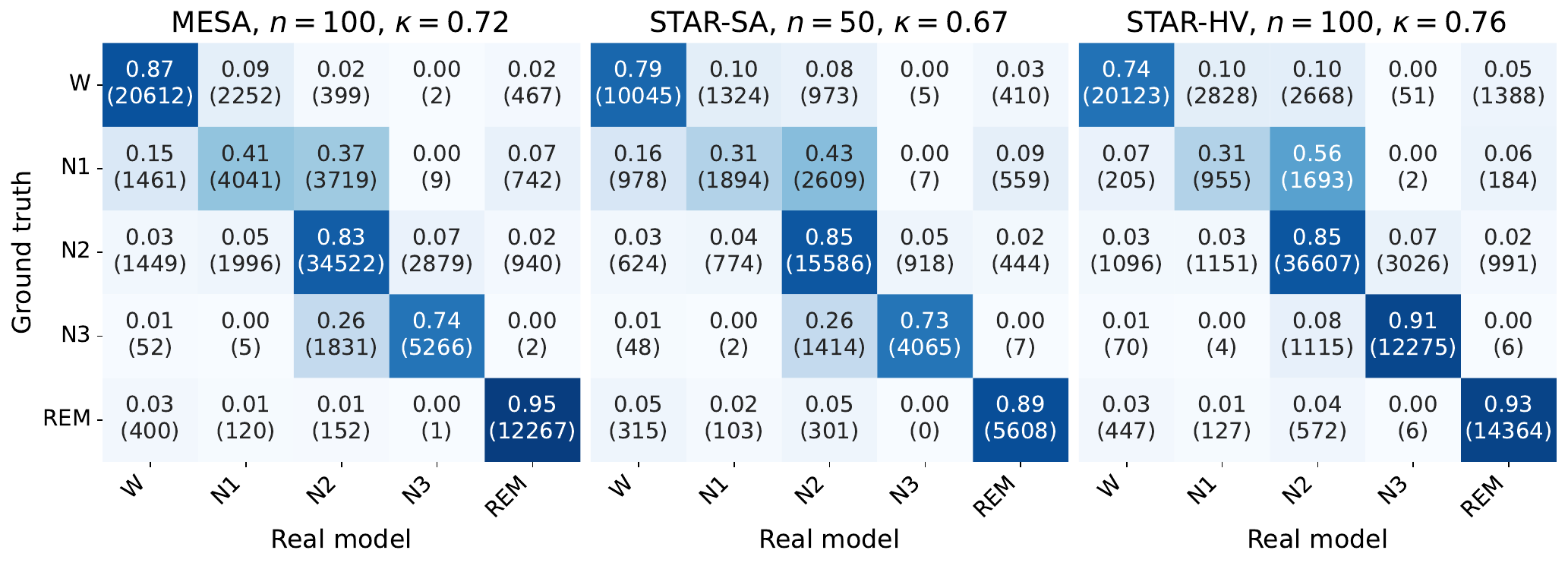}
\caption{Confusion matrix for sleep staging with 30-second resolution. W~=~wakefulness; REM~=~rapid eye movement sleep; N1, N2, and N3 = non-REM sleep stages N1--N3.}
\label{fig:sleep_staging_confmat}
\end{subfigure}

\vspace{1pt}

\begin{subfigure}[b]{\textwidth}
\centering
\includegraphics[width=0.8\textwidth]{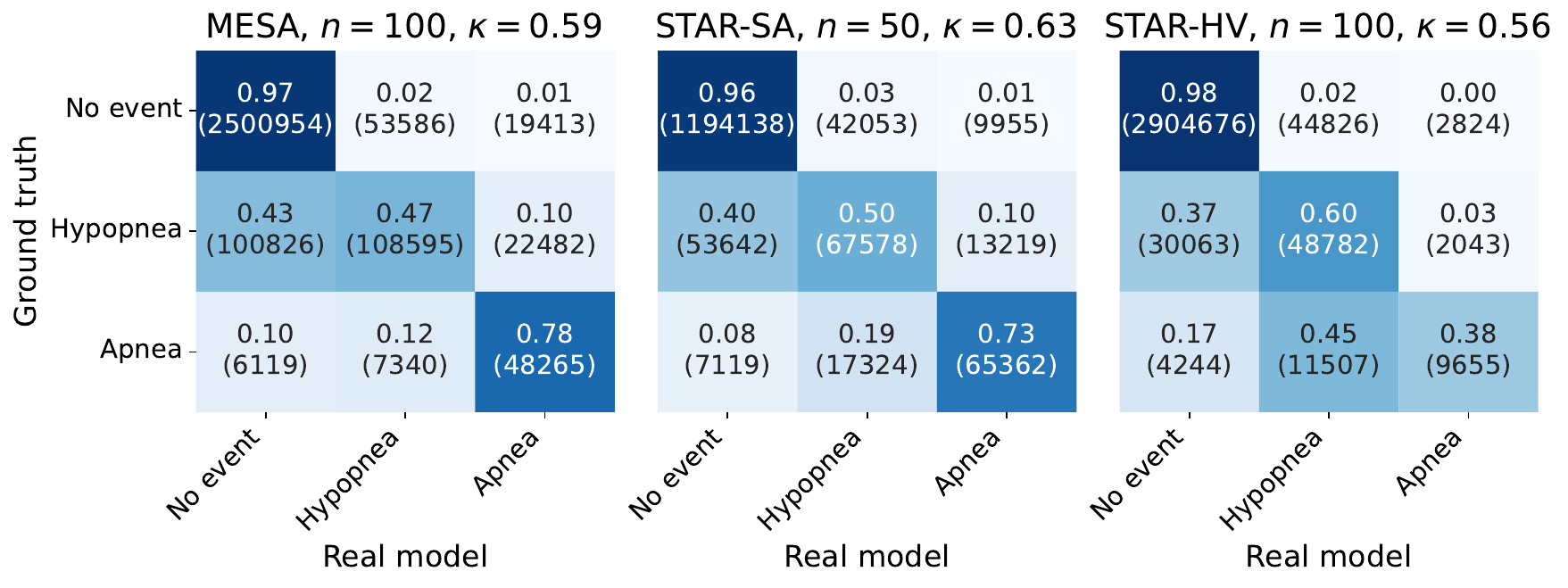}
\caption{Confusion matrix for respiratory event segmentation with 1-second resolution.}
\label{fig:respiratory_event_confmat}
\end{subfigure}

\caption{Test set confusion matrices for automatic scoring model trained with real data for (a) sleep staging and (b) apnea and hypopnea segmentation. The results are reported as fraction (segment count)}
\label{fig:scoring_confmats}
\end{figure}

The model performance was assessed with the test sets from MESA, STAR-SA, and STAR-HV. The confusion matrices for the test sets are shown in Fig.~\ref{fig:scoring_confmats}. The overall accuracy and Cohen's $\kappa$ for sleep staging was 80.3\% ($\kappa$=0.72) in the MESA test set, 75.9\% ($\kappa$=0.67) in the STAR-SA test set, and 82.7\% ($\kappa$=0.76) in the STAR-HV test set. The second-by-second accuracy and  Cohen's $\kappa$ for respiratory event detection were 92.7\% ($\kappa$=0.59) in the MESA test set, 90.3\% ($\kappa$=0.63) in the STAR-SA test set, and 96.9\% ($\kappa$=0.56) in the STAR-HV test set.

\subsection{Generative model training}

A generative adversarial network based on StyleGAN3~\cite{karrasAliasFreeGenerativeAdversarial2021}, called PSG-GAN, was trained using two million 32-minute segments randomly sampled with replacement from the STAGES PSG recordings. The PSG-GAN simultaneously generates EEG, EOG, nasal pressure, and PPG signals. To monitor the training progress and choose the final model, the Fréchet distance between the real and synthetic data was computed from the low-resolution features of the automatic scoring model, similarly to the Fréchet inception distance (FID) widely used in the context of image generation~\cite{heuselGANsTrainedTwo2017}. We named this novel metric the Fréchet U-net distance (FUD) because the low-resolution features from a U-Net type classifier~\cite{huttunenComparisonSignalCombinations2023a} were used.

\begin{figure}[htbp]
\centering
\centerline{\includegraphics[width=0.6\textwidth]{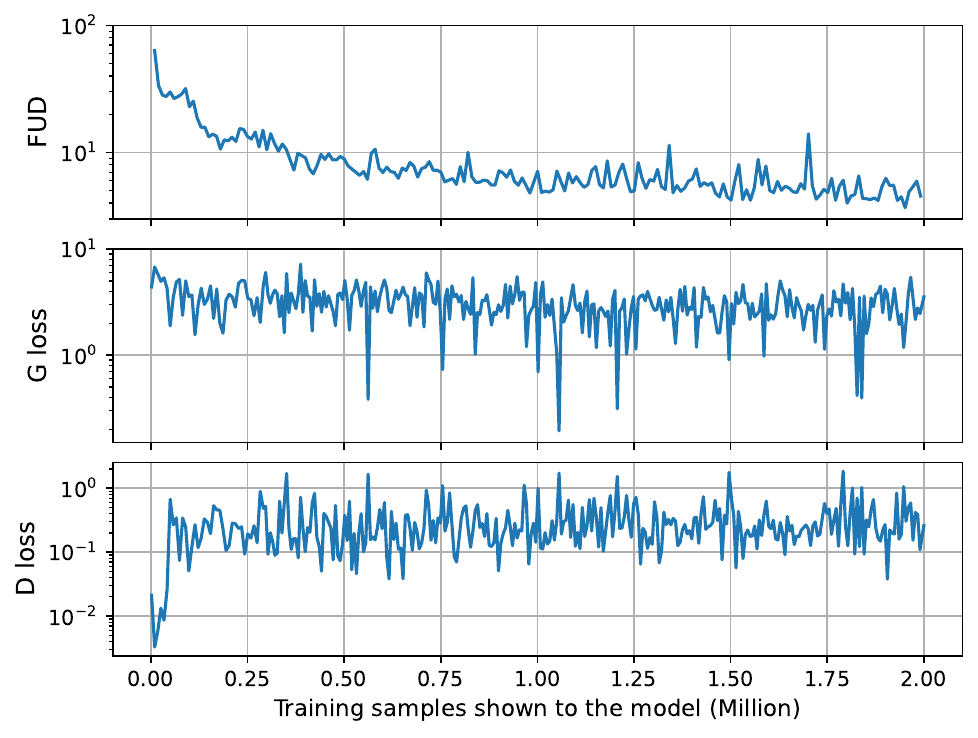}}
\caption{The progression of the Fréchet U-net distance (FUD), generator loss (G loss), and discriminator loss (D loss) during the generative model training.}
\label{fig:training_progression}
\end{figure}

The progression of the generator loss, discriminator loss, and FUD during training are shown in Fig.~\ref{fig:training_progression}. The figure shows that the generator loss reached the lowest value at around 1.05 million training samples, whereas the FUD decreased more consistently to its lowest value after 1.95 million training samples.

\subsection{Comparison of synthetic and real data}

\begin{figure}[htbp]
\centering
\begin{subfigure}[b]{1.0\textwidth}
\centering
\includegraphics[width=\textwidth]{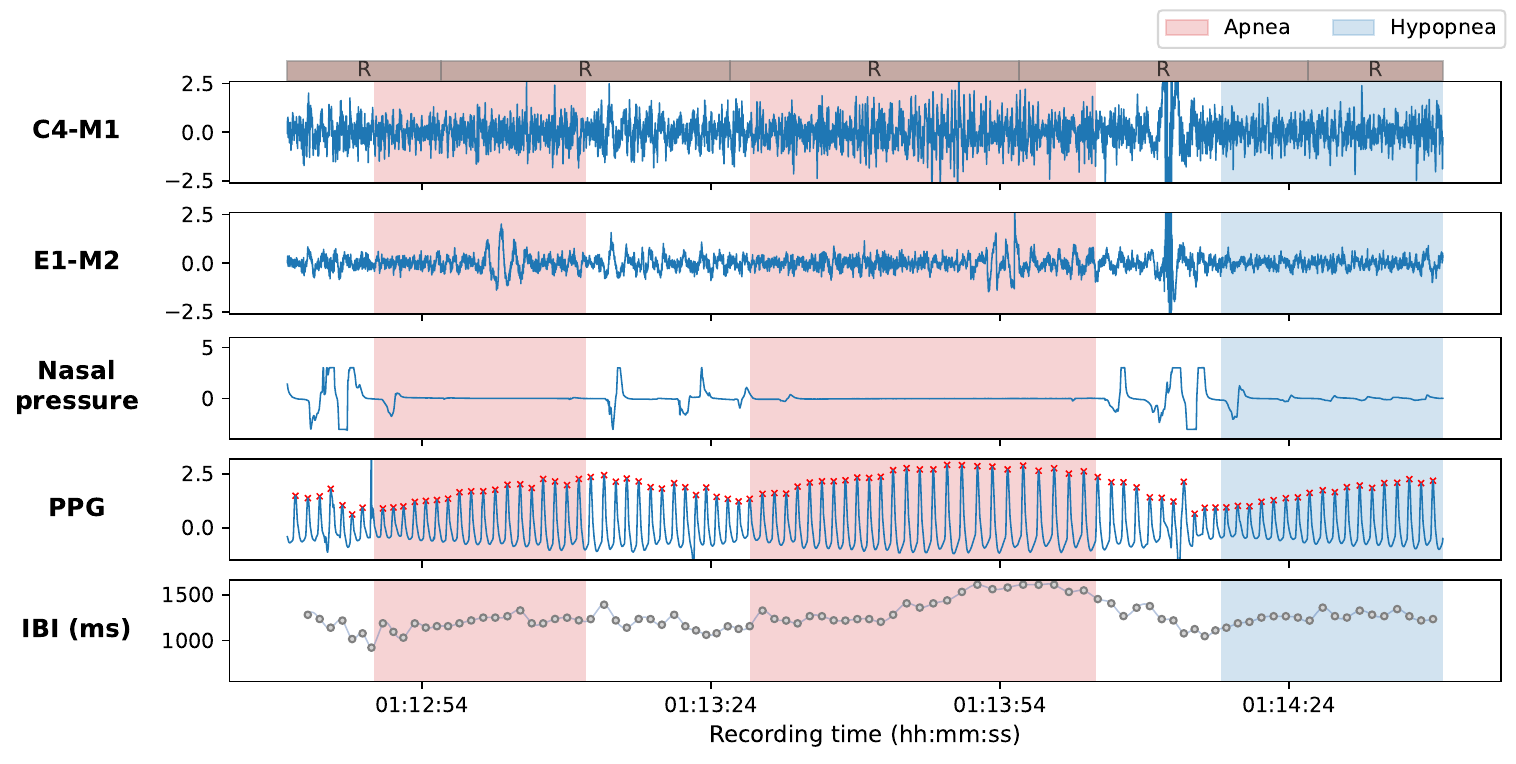}
\caption{An example of normalized real signals.}
\label{fig:example_real_signals}
\end{subfigure}


\begin{subfigure}[b]{1.0\textwidth}
\centering
\includegraphics[width=\textwidth]{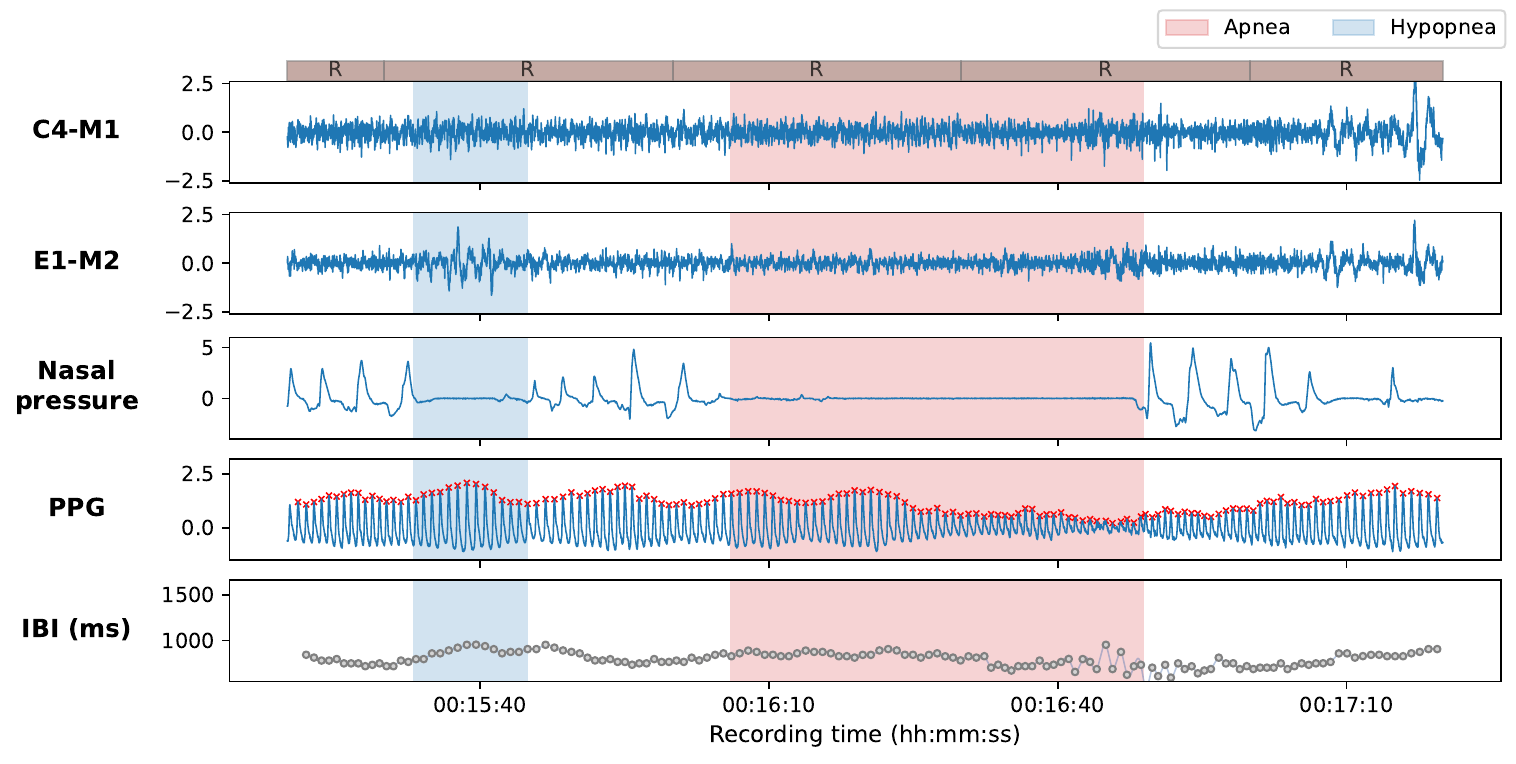}
\caption{An example of synthetic signals.}
\label{fig:example_fake_signals}
\end{subfigure}

\caption{Examples of two minutes of (a) real and (b) synthetic signals. Automatically scored apneas and hypopneas are shown on top of the signals. Inter-beat intervals (IBI) calculated from the automatically detected photoplethysmography (PPG) peaks are also shown. Due to normalization of the training data, the signal amplitude units are arbitrary. R = rapid eye movement sleep.}
\label{fig:real_vs_fake_examples}
\end{figure}

To compare the real and synthetic data, a dataset equal to the size of the STAGES dataset was generated with the PSG-GAN (19\,760 segments with a duration of 32 minutes, amounting to 10\,539 hours). Due to the high number of samples, all differences in statistics between real and synthetic data included in the results were statistically significant. Thus, statistical significance is not reported in the comparisons. Both the real and synthetic datasets were scored with the automatic scoring model introduced in Section~\ref{subsec:auto_model_real_data}. Examples of real and synthetic signals are shown in Fig.~\ref{fig:real_vs_fake_examples}. The examples were chosen to contain respiratory events in REM sleep.

\begin{figure}[htbp]
\centering
\begin{subfigure}[b]{0.8\textwidth}
\centering
\includegraphics[width=\textwidth]{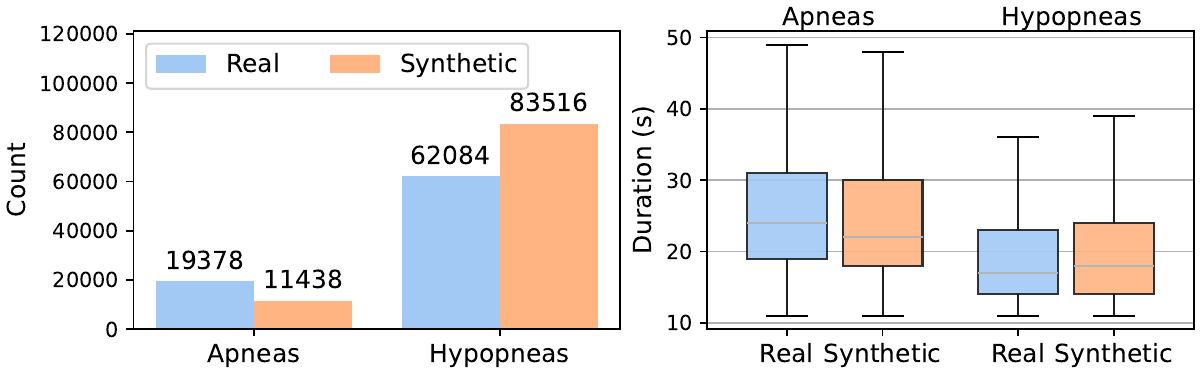}
\end{subfigure}


\begin{subfigure}[b]{0.75\textwidth}
\centering
\includegraphics[width=0.8\textwidth]{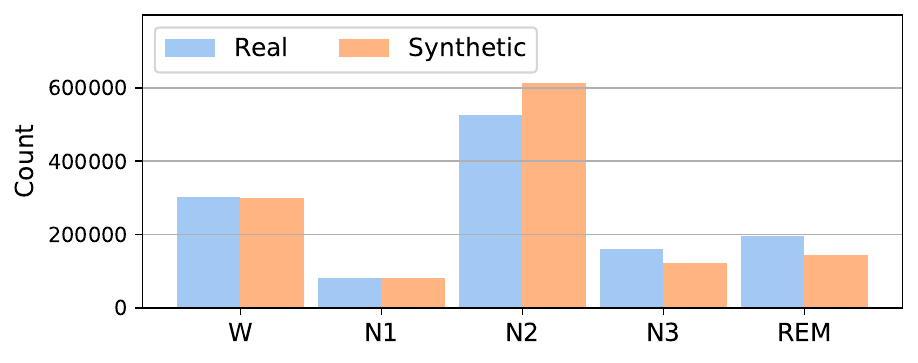}
\end{subfigure}

\caption{Counts and durations of apneas and hypopneas (upper panel) and counts of epochs by sleep stage (lower panel) based on automatic annotations. Counts are shown as bar charts. Durations are shown as box-and-whisker plots showing the medians, first and third quartiles, and $\pm$1.5 * inter-quartile range (right). W = wakefulness; REM = rapid eye movement sleep; N1, N2, and N3 = non-REM sleep stages N1--N3.}
\label{fig:stage_and_event_stats}
\end{figure}

According to the automatic scoring model, the PSG-GAN generated a greater number of hypopneas, but less apneas, compared to the STAGES data (Fig.~\ref{fig:stage_and_event_stats}). The boxplots show that the durations of real vs. synthetic events were similarly distributed for both apneas (median durations 24~s vs. 22~s), and hypopneas (median durations 17~s vs. 18~s). Total counts of automatically scored sleep stages in real and synthetic datasets are shown in Fig.~\ref{fig:stage_and_event_stats}. Overall, the distributions of sleep stages were similar. The biggest differences were that, compared to the real STAGES data, the synthetic data contained less epochs scored as N3 and REM (160\,665 vs. 122\,480 epochs of N3; 195\,248 vs. 145\,478 epochs of REM for real vs. synthetic data), and more epochs scored as N2 sleep (524\,704 vs. 614\,412 epochs for real vs. synthetic data).

\begin{figure}[htbp]
\centering
\centerline{\includegraphics[width=1.0\textwidth]{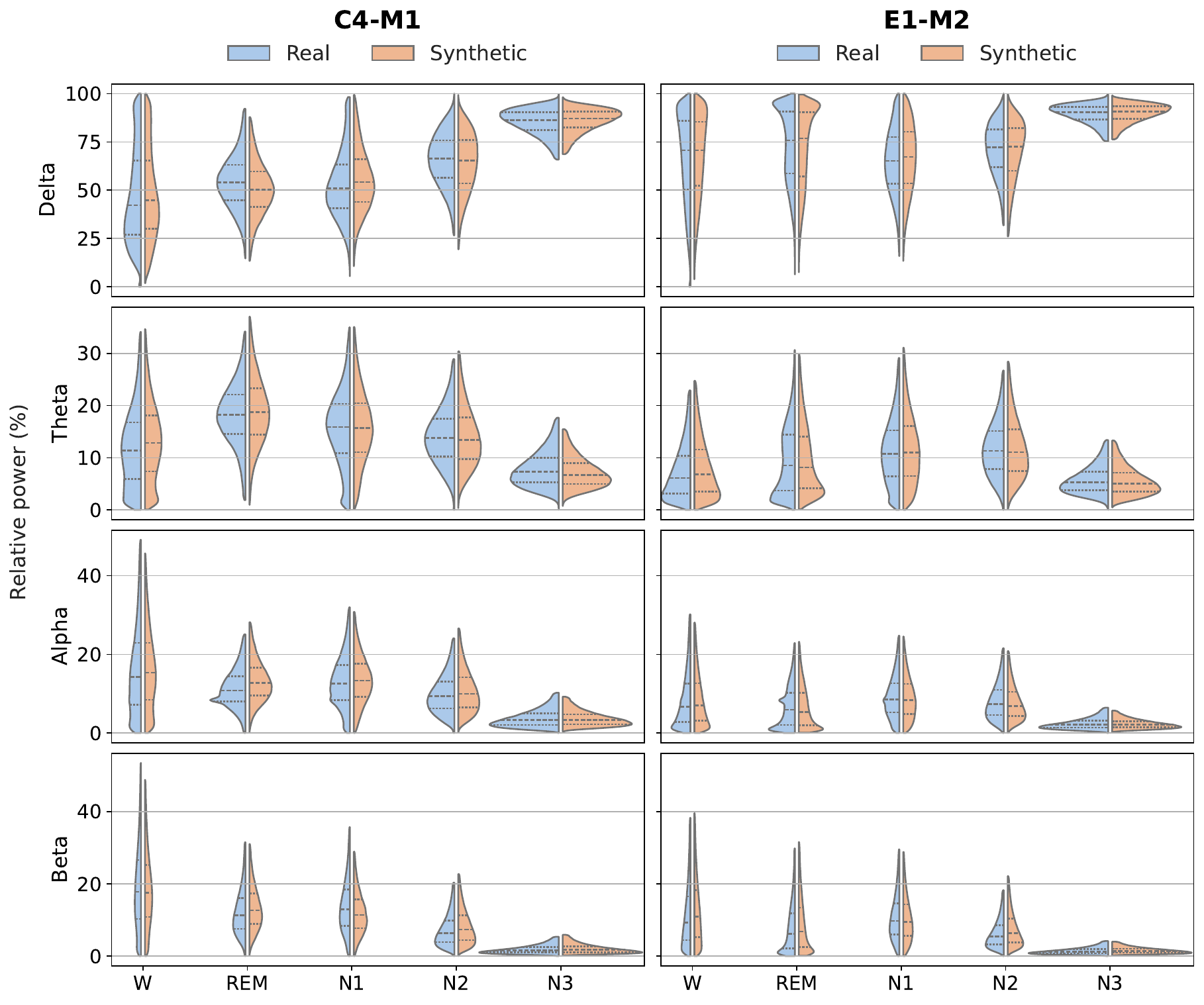}}
\caption{Violin plots of relative EEG (C4-M1) and EOG (E1-M2) powers in delta (0--4~Hz), theta (4--8~Hz), alpha (8--13~Hz), and beta (13--30~Hz) frequency bands for all 30-s epochs. The violins have been scaled so that the areas are equal in each subplot. Median, first quartile, and third quartile are shown as horizontal dashed lines. W~=~wakefulness; REM~=~rapid eye movement sleep; N1, N2, and N3 = non-REM sleep stages N1--N3.}
\label{fig:EEG_EOG_bandpowers}
\end{figure}

To compare the frequency content of real and synthetic data in the EEG and EOG signals, the relative band powers were calculated for the commonly used delta (0--4~Hz), theta (4--8~Hz), alpha (8--13~Hz), and beta (13--30~Hz) bands. The distributions of these band powers for all 30-s epochs are shown in Fig.~\ref{fig:EEG_EOG_bandpowers} as violin plots with medians, first (Q1), and third (Q3) quartiles. The figure shows that the relative band powers were similarly distributed between real and synthetic data in all sleep stages for both EEG and EOG signals.

\begin{figure}[htbp]
\centering
\centerline{\includegraphics[width=0.65\textwidth]{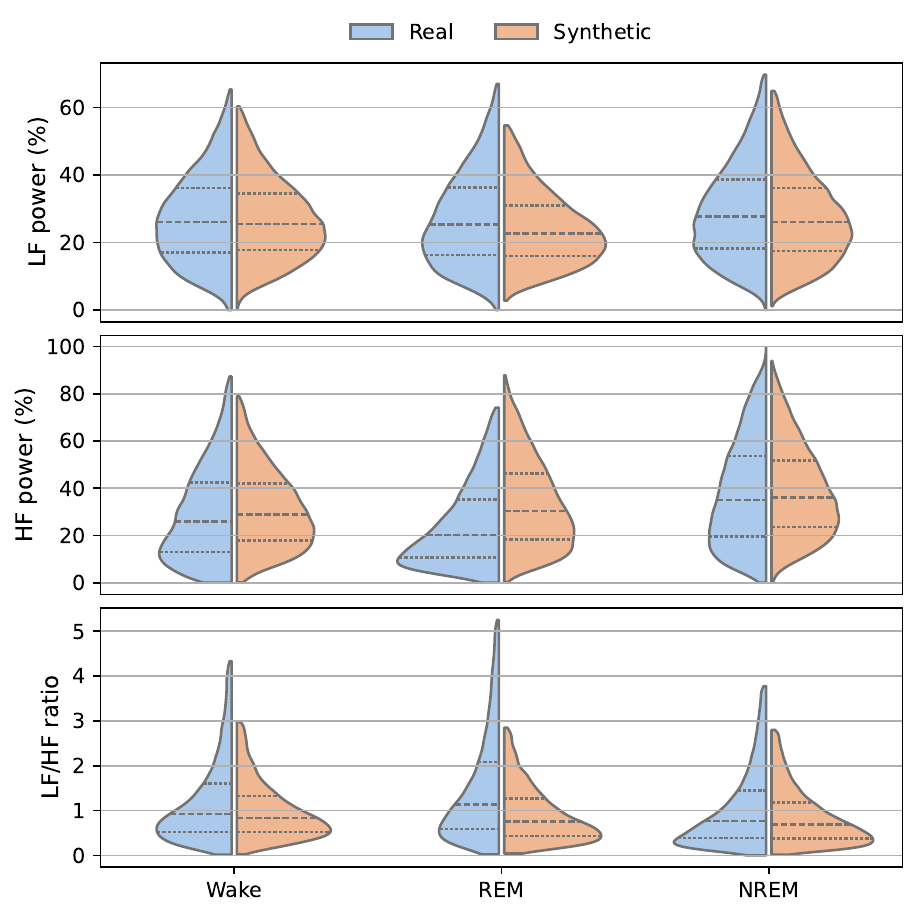}}
\caption{Violin plots of relative band powers for low frequency (LF; 0.04--0.15~Hz), high frequency (HF; 0.15--0.4~Hz), and LF/HF ratio calculated from inter-beat intervals automatically detected from the photoplethysmography signal in 5-minute segments. The band powers were calculated separately for wakefulness, rapid eye movement (REM), and non-REM (NREM) sleep.}
\label{fig:prv_bandpowers}
\end{figure}

To compare the differences in pulse rate variability (PRV) between REM and NREM sleep, relative band powers from automatically detected peaks in the PPG signal were calculated in 5-minute windows. The distributions of low-frequency (LF; 0.04--0.15~Hz) and high-frequency (HF; 0.15--0.4~Hz) relative band powers, as well as the LF/HF ratio, were plotted for wakefulness, REM, and NREM sleep (Fig.~\ref{fig:prv_bandpowers}). In wakefulness and NREM sleep, the distributions of real and synthetic data were remarkably similar. However, in REM sleep there were notable differences between real and synthetic distributions, especially in the HF band (median relative REM HF power 20.7\% vs. 30.3\% for real vs. synthetic data, respectively).

\begin{figure}[htbp]
\centering
\centerline{\includegraphics[width=1.0\textwidth]{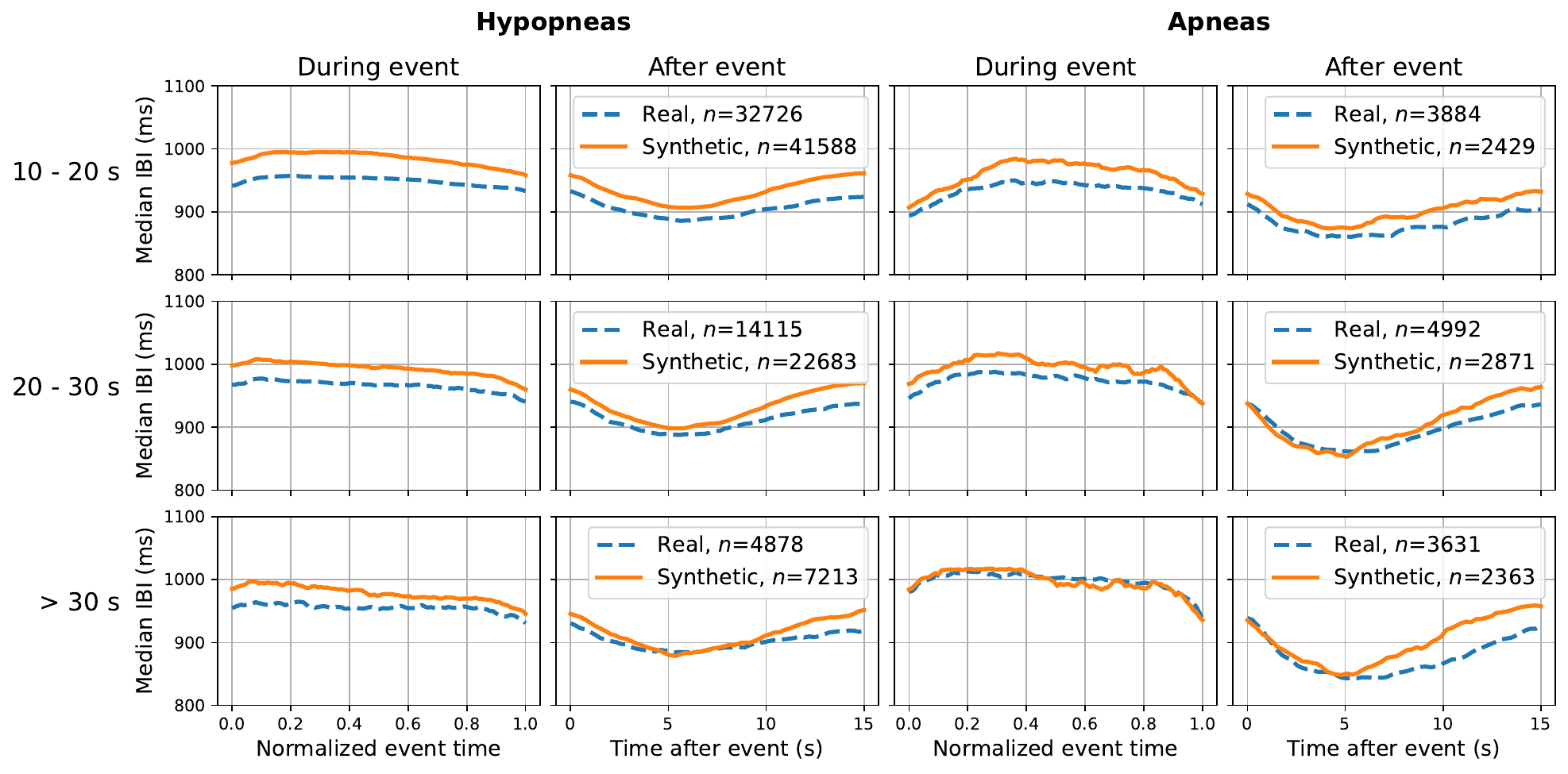}}
\caption{Median inter-beat intervals (IBI) during and after different lengths (10--20s, 20--30s, and $>$30s) of apneas and hypopneas scored with the automatic model.}
\label{fig:apnea_hypopnea_ihrs}
\end{figure}

To assess whether the PSG-GAN generated changes in heart rate related to respiratory events, median inter-beat intervals (IBI) extracted from the PPG signals during and after the automatically scored respiratory events were compared between real and synthetic data (Fig.~\ref{fig:apnea_hypopnea_ihrs}). Overall, the synthetic data showed decrease in the IBIs after respiratory events similarly to the real data. However, the median IBI was consistently higher in the synthetic data during and after the event regardless of event duration and type, except for apneas longer than 30s, where the medians between real and synthetic IBIs were similar during the events.

\subsection{Automatic scoring model trained with synthetic data}

To assess the applicability of the synthetic signals for AI training, an automatic scoring model was trained from scratch with the automatically scored synthetic data used in the previous section. The synthetic dataset was split to training ($n$=19060), validation ($n$=200), and test ($n$=500) sets. The synthetic scoring model was additionally validated and tested with the validation and test sets from the MESA, STAR-SA, and STAR-HV datasets.

\begin{figure}[htbp]
\centering

\begin{subfigure}[b]{\textwidth}
\centering
\includegraphics[width=\textwidth]{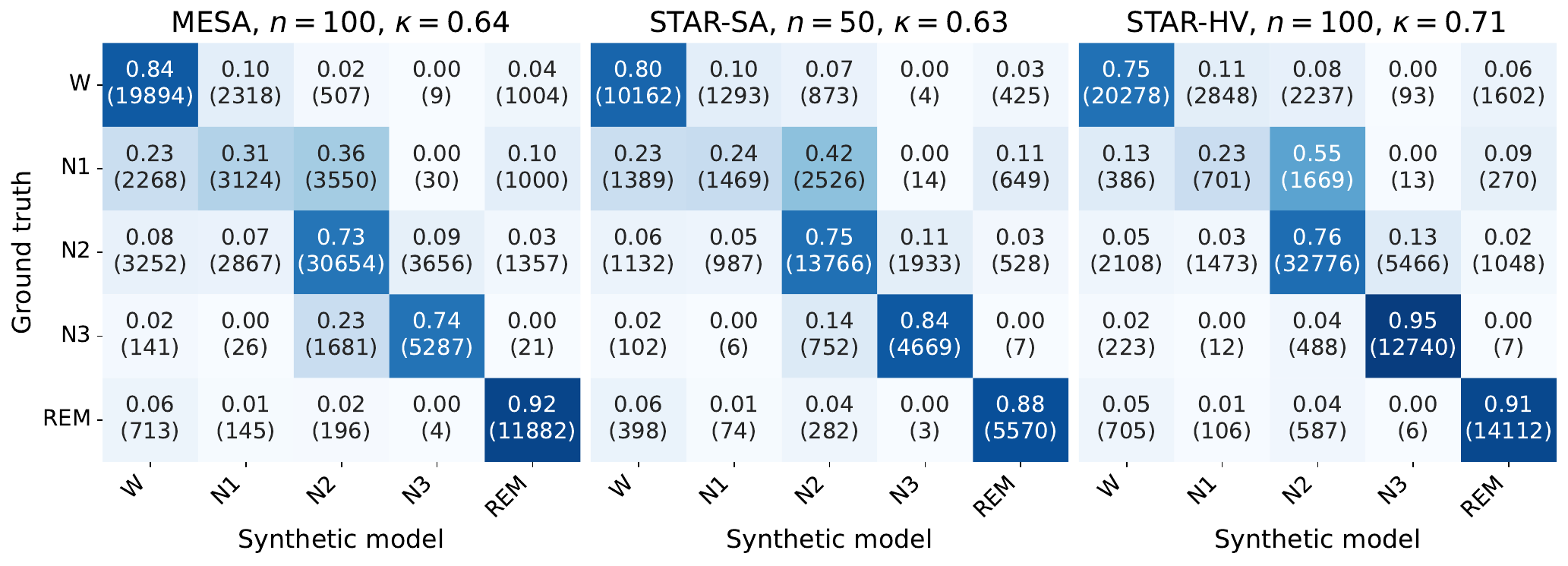}
\caption{Confusion matrices for sleep staging with 30-second resolution. W~=~wakefulness; REM~=~rapid eye movement sleep; N1, N2, and N3 = non-REM sleep stages N1--N3.}
\label{fig:test_hg_confmats}
\end{subfigure}

\vspace{1pt}

\begin{subfigure}[b]{\textwidth}
\centering
\includegraphics[width=0.8\textwidth]{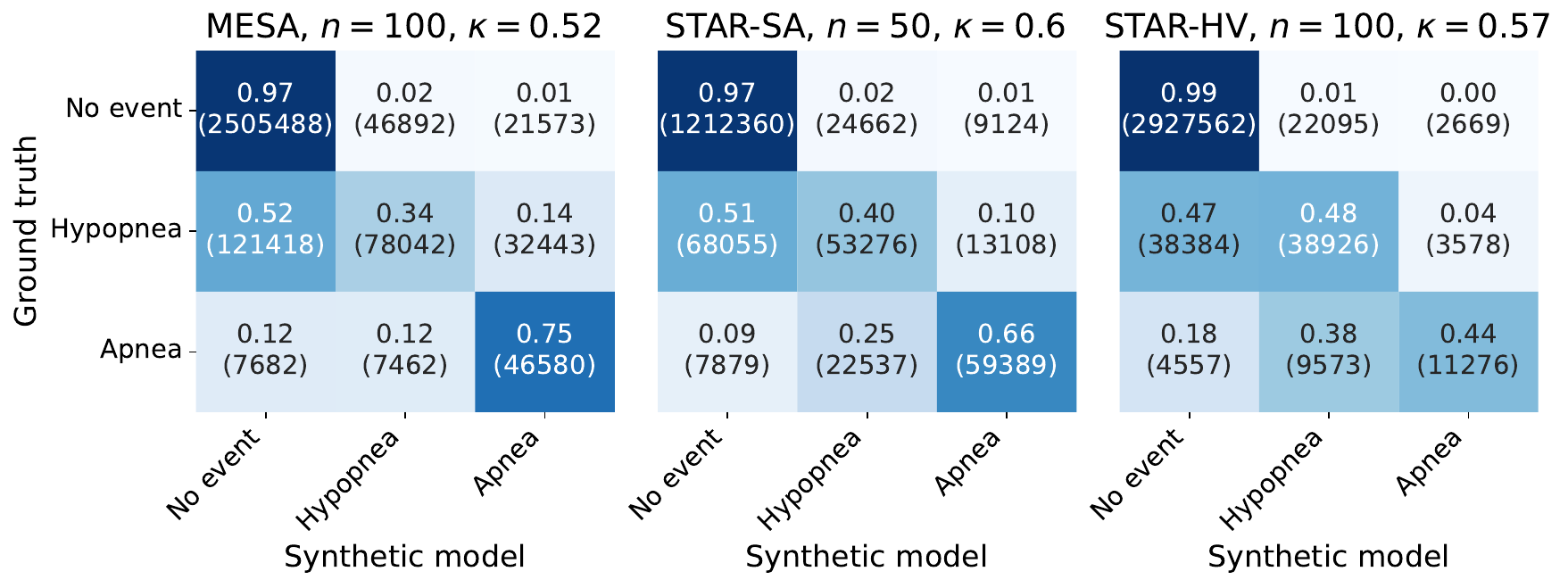}
\caption{Confusion matrices for respiratory event segmentation with 1-second resolution.}
\label{fig:test_events_confmats}
\end{subfigure}

\caption{Confusion matrices on different test sets for models trained with synthetic data for (a) sleep stages and (b) apneas and hypopneas. The results are reported as fraction (segment count).}
\label{fig:test_confmats}
\end{figure}

The confusion matrices for ground truth vs. synthetic scoring model are shown in Fig.~\ref{fig:test_confmats}. The performance of the synthetic scoring model was good for sleep staging: the accuracy and Cohen's $\kappa$ values were 74.1\% ($\kappa$=0.64) in the MESA test set, 72.7\% ($\kappa$=0.63) in the STAR-SA test set, and 79.0\% ($\kappa$=0.71) in the STAR-HV test set. In respiratory event segmentation with 1-second resolution, the synthetic model's overall accuracy and Cohen's $\kappa$ were 91.7\% ($\kappa$=0.52) in the MESA test set, 90.1\% ($\kappa$=0.60) in the STAR-SA test set, and 97.4\% ($\kappa$=0.57) in the STAR-HV test set.

\clearpage
\section{Discussion}\label{sec:discussion}

In this study, we developed the PSG-GAN, a generative adversarial network designed to synthesize polysomnography signals and their OSA-related interactions. The PSG-GAN was based on the StyleGAN3 architecture, and was trained to simultaneously synthesize EEG, EOG, nasal pressure, and PPG signals. In line with the first hypothesis, the synthetic signals appeared realistic (Fig.~\ref{fig:real_vs_fake_examples}), and EEG and EOG signals were distributed similarly to the real training data in terms of relative powers on the delta, theta, alpha, and beta frequency bands (Fig.~\ref{fig:EEG_EOG_bandpowers}). The inter-beat intervals extracted from the automatically detected PPG peaks followed similar trends during and after respiratory events in both real and synthetic data (Fig.~\ref{fig:apnea_hypopnea_ihrs}), which supported the second hypothesis that the PSG-GAN would generate OSA-related signal interactions. Automatic scoring models trained solely on synthetic data reached good performance on real test data (Fig.~\ref{fig:test_confmats}), which was in line with the third hypothesis. This demonstrates the potential of using synthetic data for algorithm development in the diagnostics of sleep disorders.

Apneas and hypopneas have been shown to cause changes in the instantaneous heart rate, and the magnitude of these changes is modulated by the duration and type of the respiratory events~\cite{hietakosteLongerApneasHypopneas2020}. The PSG-GAN was capable of generating similar changes seen in the inter-beat intervals extracted from the PPG signal during and after respiratory events (Fig.~\ref{fig:apnea_hypopnea_ihrs}). In both real and synthetic signals, the heart rate tends to slow down at the beginning of an obstruction, and then increase slowly towards the end of the event. After the obstruction, heart rate increases rapidly and then starts to decrease again during the 15-second post-event period. Moreover, longer and complete obstructions cause a greater change in the heart rate during and after the event~\cite{hietakosteLongerApneasHypopneas2020}. However, the PSG-GAN did not fully learn to capture the interactions between the neuro-cardio-respiratory systems represented by the shifts in the LF and HF band powers of pulse rate variability between REM and NREM sleep (Fig.~\ref{fig:prv_bandpowers}). In real subjects the high-frequency activity is notably changed during REM sleep, while the changes are not as clear in the synthetic data (Fig.~\ref{fig:prv_bandpowers}).

The StyleGAN3 architecture has several key aspects that are suitable for synthesizing long segments of PSG signals. First, the synthesis part of the generator starts from random Fourier features which consist of randomly sampled sine waves with frequencies from 0 to 0.1~Hz. Secondly, the new frequency content introduced by the synthesis layers is heavily limited, especially in the earlier layers of the synthesis network (Table~\ref{tab:network_params}). This enforces the generator to initially focus on the low-frequency features that are mutual to the signals, even when the output signals have vastly different distributions of spectral power. We hypothesize that these low-frequency features contain information on sleep-related cyclical activity such as sleep cycles and obstructed breathing patterns. The analysis and synthesis of these low-frequency features could be an interesting direction for future research.

With GANs, it is desirable that smooth interpolations between points in the latent space lead to smooth transformations in the generated signals. With PSG-GAN, this can be seen in the accompanied Supplementary Video 1, which shows the progression of the signals while interpolating between randomly sampled points in the latent space. The signals in the video change smoothly without sudden jumps, and show variation between different sleep stages (e.g. in delta waves and eye movements), as well as between normal and disordered breathing. This suggests that the PSG-GAN did not suffer from mode collapse; it did not simply memorize the training samples, instead producing a continuous stream of new data. This could be especially useful when applying the PSG-GAN to data augmentation and semi-supervised learning.

One limitation of this study was that nasal pressure was the only respiratory signal utilized to reduce the number of signals used in the deep learning models. However, the nasal pressure signals were flat or noisy for parts of the training data, which was propagated to the synthetic data. The amplification of these artifacts in synthetic nasal pressure signals may be one reason for the reduced performance of respiratory event segmentation with the synthetic model. In OSA diagnostics, the respiratory events are scored using multiple respiratory signals, including oronasal thermistor and respiratory belts. Thus, with full PSG measurement setup, the respiratory events can be scored even if some of the respiratory signals are unusable. Nevertheless, all nasal pressure signals were included in the PSG-GAN training to maximize the amount of training data, since these types of artifacts are very common in nasal pressure monitoring.

Another limitation was the quality of manual annotations in the datasets. For example, there were no hypopneas scored in eight out of thirteen STAGES sites. Moreover, in some of the STAGES sites, all respiratory events were located in the beginning of a sleep staging epoch, not when they actually started. There were also inconsistencies in manual sleep staging in some of the sites. Thus, we chose to use automatic annotations when assessing the quality of the synthetic signals. The MESA dataset that was used to train the automatic scoring model consists of elderly subjects with high OSA prevalence, which may be sub-optimal when scoring the younger and healthier STAGES population. Still, the automatic scoring models trained with automatically scored synthetic data performed well when compared to manually scored test data, which supports the use of the automatic scoring model (Fig.~\ref{fig:test_confmats}).

Alternative training methods such as feature matching~\cite{salimansImprovedTechniquesTraining2016a} could improve the quality of synthetic data. However, feature matching involves computing the difference between real and synthetic features, and using real examples in the generator training step would make it easier for the generator to memorize the training data. Instead, during the training of PSG-GAN, the generator weights were adjusted based only on synthetic data. Moreover, conditioning the PSG-GAN on the annotations, such as with conditional GANs~\cite{mirzaConditionalGenerativeAdversarial2014b} or auxiliary classifier GANs~\cite{odenaConditionalImageSynthesis2017a}, could improve the quality of the generated data. However, the goal of this study was to utilize the signals without annotations, and making the GAN conditional would defeat the purpose.

One interesting future direction would be the inversion of the generator and performing, for example, clustering of real PSG signals in the latent space. Since the initial random Fourier features of the synthesis network are static signals with fixed frequencies, the generator could be used to model signals of varying length, although in this study the length was fixed to 32 minutes. Training the PSG-GAN with variable-length segments could enable the analysis of full-night PSG recordings, that have intractable distribution, in the latent space, which is normally distributed and where the number of dimensions is much lower and fixed. Another future improvement would be to increase the sampling rate to include the EEG gamma band, consisting of frequencies up to 100~Hz. We hypothesize that the PSG-GAN could also be used as an anonymization technique to create large-scale, open access synthetic datasets based on heavily regulated clinical data. However, quantifying the level of anonymity warrants further studies.

In conclusion, the PSG-GAN is capable of generating realistic PSG signals with greatly varying characteristics. The generator learned to model signal interactions characteristic to OSA, such as changes in heart rate during breathing obstructions. Automatic scoring models trained with the synthetic signals generalized well to real data. Commercialization of health care applications utilizing clinical PSG data is constrained by data regulation. Thus, it is crucial to develop new methods to generate synthetic PSG data that could substitute the sensitive patient data in AI-based health care applications of the future.

%
%

\section{Methods}\label{sec:methods}

\subsection{Data selection and preprocessing}\label{subsec:data_preprocessing}

To train the models, one EEG channel (C4-M1), one EOG channel (E1-M2), nasal pressure, and PPG signals were used. This combination was selected to enable accurate classification of both sleep stages and respiratory events. The full-night recordings were resampled to 64~Hz to reduce computational requirements while preserving the frequencies up to 30~Hz, where the EEG beta frequency band ends. The EEG and EOG signals were then high-pass filtered with a cutoff of 0.3~Hz as recommended by the AASM~\cite{berryAASMManualScoring2018}. The downsampled and filtered recordings were cropped between the lights off and lights on timestamps when provided with the recordings. Finally, the cropped signals were normalized to zero mean and unit variance.

The Multi-Ethnic Study of Atherosclerosis (MESA) dataset consisted of 2\,052 full-night PSG recordings collected at multiple sites during the years 2010--2013~\cite{chenRacialEthnicDifferences2015}. Institutional review boards approved the study at each site, and all participants accepted a written informed consent. The data was recorded using the Compumedics Somte System (Compumedics Ltd., Abbotsville, Australia). Sleep stages and respiratory events were manually scored at the recording site conforming to the MESA polysomnography scoring manual available at the NSRR sleep data website (\url{https://sleepdata.org/datasets/mesa}). After excluding recordings with less than 1 hour of sleep, and recordings with corrupted or missing signals, the remaining 2052 recordings were used to train, validate, and test the automatic scoring model.

The Stanford Technology Analytics and Genomics in Sleep (STAGES)~\cite{zhangNationalSleepResearch2018} dataset was used to train the PSG-GAN. After excluding multiple sleep latency tests, recordings with less than 1 hour of sleep, and recordings with corrupted or missing signals, the remaining 1\,402 recordings were used to train the model. The STAGES dataset contains PSGs recorded at 13 sites from five different sleep centers using different measurement setups. Details on the sites and recording setups can be found on the NSRR sleep data website (\url{https://sleepdata.org/datasets/stages}). The PSG recordings were manually scored at the recording sites. However, there were inconsistencies in the manual annotations from the different sites. For this reason, all STAGES PSG recordings were rescored with the automatic scoring model for the comparison of real vs. synthetic data.

The STAR-SA dataset was collected during the years 2024 and 2025 at the SmartSleep laboratory, University of Eastern Finland (Kuopio, Finland). The PSG recordings were conducted with the Compumedics Grael 4K PSG:EEG amplifier and Compumedics ProFusion PSG 5 software (Compumedics Ltd., Australia). The participants were recently diagnosed with OSA, and the recordings were obtained between the diagnosis and start of continuous positive airway pressure treatment. The recordings were initially scored automatically with a previously developed model~\cite{huttunenComparisonSignalCombinations2023a}, and the automatic annotations were carefully adjusted by sleep technicians according to the AASM guidelines. The collection and use of the PSG recordings were approved by the Regional Medical Research Ethics Committee of Eastern Finland Collaborative Area (17/2023).

The STAR-HV dataset was collected during the years 2023--2025 at the SmartSleep laboratory, University of Eastern Finland (Kuopio, Finland). The PSG recordings were conducted with the Compumedics Grael 4K PSG:EEG amplifier and Compumedics ProFusion PSG 5 software (Compumedics Ltd., Australia). The participants were interviewed prior to obtaining the PSG recordings, and did not report any sleep disorders. The sleep stages and respiratory events were automatically scored with a previously developed model~\cite{huttunenComparisonSignalCombinations2023a}. The collection and use of the recordings was approved by the Regional Medical Research Committee of the Wellbeing Services County of North Savo (66/2023 and 135/2023).

\subsection{Automatic scoring}
\label{subsec:automatic_scoring}

The automatic scoring model used to compute the Fréchet U-net distance, and to score the STAGES and synthetic datasets, was based on a previously developed model~\cite{huttunenComparisonSignalCombinations2023a}. The model architecture consists of fully convolutional encoder and decoder, with an atrous spatial pyramid pooling (ASPP) block that captures features on multiple resolutions, in between. The model can be trained with any number of input signals and output annotations, and supports different temporal resolutions of the annotations. The only change to the original scoring model~\cite{huttunenComparisonSignalCombinations2023a} (apart from different combinations of input signals and output annotations) was that the number of output features from the ASPP block was reduced from 256 to 48 to speed up the calculation of the FUD.

To train the scoring model, the MESA dataset was randomly split into training ($\textit{n}=1852$), validation ($\textit{n}=100$), and test ($\textit{n}=100$) sets. In addition, STAR-SA and STAR-HV datasets were used for validation (STAR-SA $n$=20; STAR-HV $n$=156) and testing (STAR-SA $n$=50; STAR-HV $n$=100). The model was then trained with EEG, EOG, nasal pressure, and PPG signals using the Yogi optimizer~\cite{zaheerAdaptiveMethodsNonconvex2018}, with default hyperparameters $\beta_1 = 0.9$, $\beta_2 = 0.999$, and $\epsilon = 0.001$. The model was trained for 150 epochs with batch size of 8. On each epoch, the model was trained with randomly sampled one hour segment from each patient in the training set. The learning rate was initially set to 0.005, then increased linearly to 0.01 in 40 epochs, then decreased linearly back to 0.0025 in 40 epochs, and finally decreased linearly to 0.001 in 20 epochs.

Focal loss~\cite{linFocalLossDense2017} with $\gamma$=1.0 was used to train the models. Moreover, to take into account insufficient representation of apneas in the MESA annotations, the class-wise loss for apneas was multiplied by three. In addition, the MESA scoring rules lead to scoring less N3 and more N2 compared to the AASM scoring rules. Thus, the class-wise losses for N3 and N2 were multiplied by 2.0 and 0.7, respectively. The lowest mean loss of all validation sets (MESA, STAR-SA, and STAR-HV) was used to select the final model. All other details in training were kept the same as in~\cite{huttunenComparisonSignalCombinations2023a}.

The same architecture with the same hyperparameters was also trained with the synthetic dataset created to compare the synthetic and real data. The synthetic data was split into training ($n$=19\,060), validation ($n$=200), and test ($n$=500) sets. In addition, the MESA, STAR-SA, and STAR-HV datasets were used for validation and testing with the same splits as the automatic model trained with real data. The focal losses from all validation sets were averaged to choose the final model.

\subsection{Generative modeling}

\begin{figure}[ht]
\centering
\begin{subfigure}[b]{0.63\textwidth}
\centering
\centerline{\includegraphics[width=\textwidth]{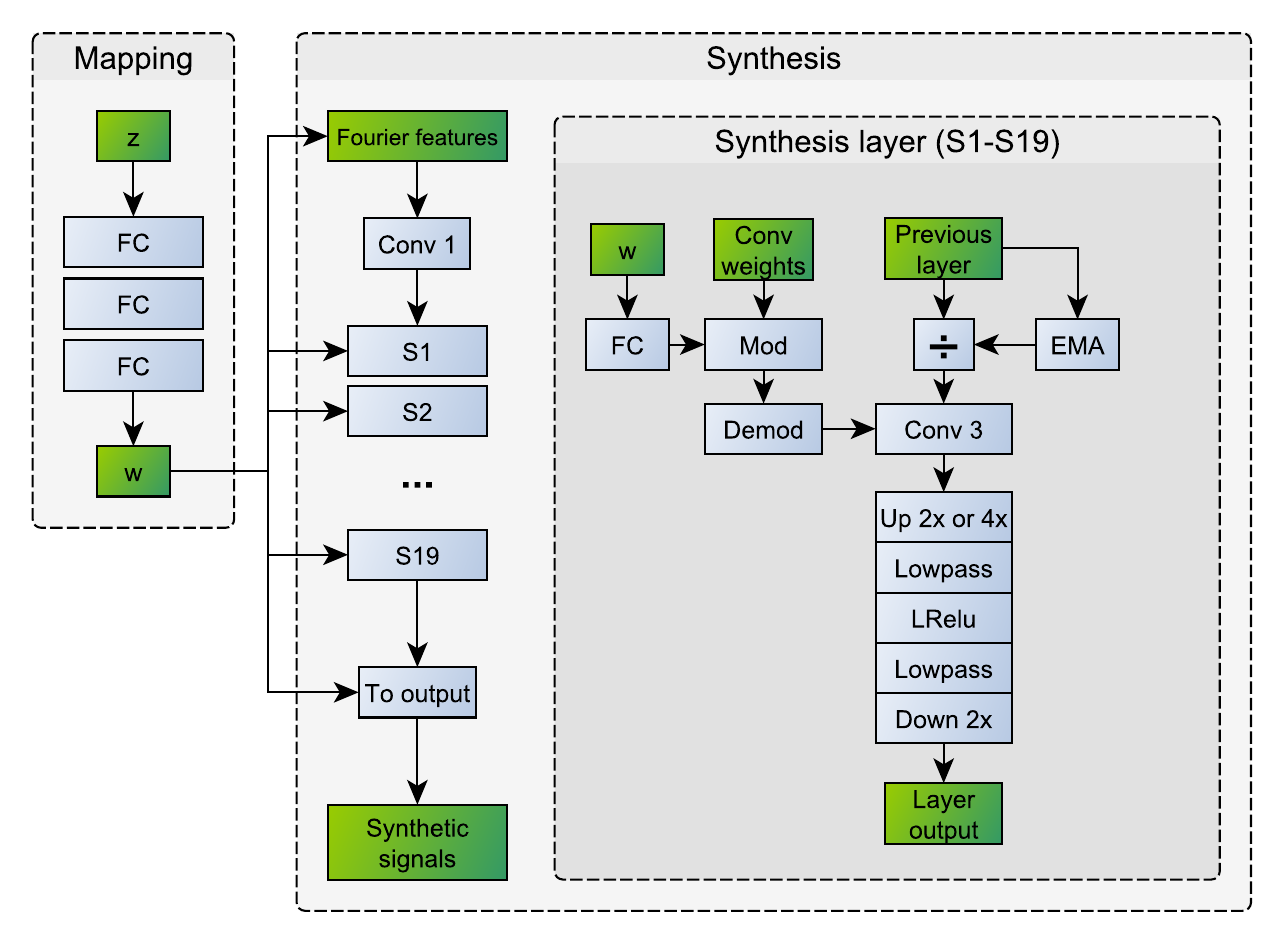}}
\caption{The generator architecture.}
\label{fig:generator}
\end{subfigure}\quad
\begin{subfigure}[b]{0.34\textwidth}
\centering
\centerline{\includegraphics[width=\textwidth]{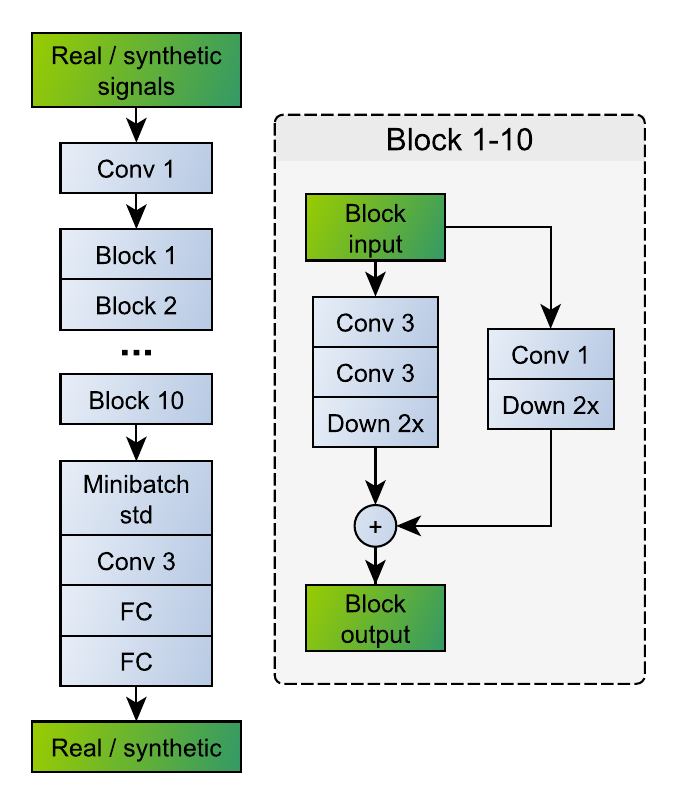}}
\caption{The discriminator architecture.}
\label{fig:discriminator}
\end{subfigure}

\caption{The architecture of the PSG-GAN. The output layer of the synthesis network is similar to the synthesis layers, but without any up/downsampling, filtering, or activation. FC = fully connected; Conv 1 or 3 = convolution with kernel size 1 or 3; EMA = exponential moving average; LRelu = leaky rectified linear unit; std = standard deviation; Mod/Demod = modulation and demodulation from the StyleGAN2~\cite{karrasAnalyzingImprovingImage2020}.}
\label{fig:gan_architecture}
\end{figure}

The PSG-GAN is a generative adversarial network based on the StyleGAN3 architecture~\cite{karrasAliasFreeGenerativeAdversarial2021}. Fig.~\ref{fig:gan_architecture} shows a schematic illustration of the model architecture. The generator consists of mapping and synthesis networks. The mapping network takes an input vector $z$ of length 512 randomly sampled from the standard normal distribution. The vector $z$ is then passed through a fully connected network with three layers each having 512 features. The resulting vector $w$ is then passed to each layer of the synthesis network. The synthesis starts from random Fourier features~\cite{karrasAliasFreeGenerativeAdversarial2021}. A learned mapping from $w$ to a phase shift is applied to the Fourier features, and the shifted features are then passed through the synthesis layers. Each synthesis layer applies a convolution modulated by a learned mapping from $w$, called a \textit{style} in the StyleGAN literature. The convolution weights are additionally normalized with the exponential moving average of the layer inputs. The modulated convolution is followed by upsampling and lowpass filtering of the signals, applying the leaky rectified linear unit (LReLU) activation to the upsampled representation, and then lowpass filtering and downsampling the signals back to desired sampling rate. In the output layer, only the modulated convolution is applied without any resampling or filtering. More details can be found in the original StyleGAN3 paper~\cite{karrasAliasFreeGenerativeAdversarial2021}.

StyleGAN3 uses the discriminator architecture from StyleGAN2~\cite{karrasAnalyzingImprovingImage2020}, which consists of residual blocks that each apply two convolutions followed by activations to the inputs (Fig.~\ref{fig:discriminator}). The outputs of the block are downsampled by a factor of two in the temporal dimension. After the discriminator blocks, a minibatch standard deviation layer~\cite{karrasProgressiveGrowingGANs2018} is applied to enable the discriminator to assess the variability between the samples. Finally, fully connected layers are applied to the features. The output is a scalar that expresses whether the discriminator inputs are real or synthetic.

The original StyleGAN3 was slightly modified for this study. First, all 2D operations, such as convolution and filtering, were replaced by their 1D counterparts. Secondly, instead of the rotation and translation of the Fourier features, only a phase shift was applied to the 1D features. Thirdly, the number of synthesis layers, as well as their sampling, cutoff, and transition band half width frequencies were modified to work with the PSG signals (Table~\ref{tab:network_params}). In the PSG-GAN, the initial Fourier features have frequencies up to 0.1~Hz in 0.25~Hz sampling rate. The synthesis network layers then gradually increase the sampling rate up to 64~Hz. The cutoff and transition band half width sum up to the Nyquist frequency, meaning that all frequencies above the Nyquist frequency are on the stopband, except for the last three layers, in which the cutoff frequency is set to Nyquist frequency to enable generation of high-frequency details.

\begin{table*}
\centering
\caption{Network parameters.}
\label{tab:network_params}
\begin{threeparttable}
\begin{tabular}[t]{lrrrr}
\toprule
\multicolumn{5}{c}{Synthesis network} \\
\midrule
 & Features & $F_s$ & $F_c$ & $F_{h}$ \\
Layer &  &  &  &  \\
\midrule
1 & 256 & 0.25 & 0.10 & 0.02 \\
2 & 256 & 0.50 & 0.14 & 0.11 \\
3 & 256 & 0.50 & 0.20 & 0.05 \\
4 & 256 & 1.0 & 0.28 & 0.22 \\
5 & 256 & 1.0 & 0.39 & 0.11 \\
6 & 256 & 2.0 & 0.55 & 0.45 \\
7 & 256 & 2.0 & 0.77 & 0.23 \\
8 & 256 & 4.0 & 1.08 & 0.92 \\
9 & 256 & 4.0 & 1.51 & 0.49 \\
10 & 242 & 8.0 & 2.12 & 1.88 \\
11 & 172 & 8.0 & 2.98 & 1.02 \\
12 & 123 & 16.0 & 4.18 & 3.82 \\
13 & 87 & 16.0 & 5.87 & 2.13 \\
14 & 62 & 32.0 & 8.24 & 7.76 \\
15 & 44 & 32.0 & 11.56 & 4.44 \\
16 & 32 & 64.0 & 16.23 & 15.77 \\
17 & 22 & 64.0 & 22.79 & 9.21 \\
18 & 16 & 64.0 & 32.0 & 3.20 \\
19 & 16 & 64.0 & 32.0 & 3.20 \\
20 & 4 & 64.0 & 32.0 & 3.20 \\
\bottomrule
\end{tabular}

\begin{tablenotes}
\item $F_s$ = sampling rate (hertz)
\item $F_c$ = cutoff frequency (hertz)
\item $F_{h}$ = transition band half width (hertz)
\end{tablenotes}
\end{threeparttable}\quad
\begin{threeparttable}
\begin{tabular}[t]{lrr}
\toprule
\multicolumn{3}{c}{Discriminator} \\
\midrule
 & Features & $F_s$ \\
Block &  &  \\
\midrule
1 & 16 & 64.0 \\
2 & 32 & 32.0 \\
3 & 64 & 16.0 \\
4 & 128 & 8.0 \\
5 & 256 & 4.0 \\
6 & 256 & 2.0 \\
7 & 256 & 1.0 \\
8 & 256 & 0.50 \\
9 & 256 & 0.25 \\
10 & 256 & 0.125 \\
\\
\\
\\
\\
\\
\\
\\
\\
\\
\\
\end{tabular}

\end{threeparttable}
\end{table*}

The PSG-GAN was trained to generate EEG, EOG, nasal pressure, and PPG signals at a 64~Hz sampling rate with 2 million 32-minute segments randomly sampled from the STAGES dataset with a batch size of 8. At each step, the discriminator parameters were first adjusted using a batch of both real and synthetic data. The generator parameters were then adjusted using a different batch of synthetic data only. The Adam optimizer was used for both the generator and discriminator, with $\beta_1 = 0$ and $\beta_2 = 0.99$. The learning rate was kept at 0.001 during the first 125\,000 steps, and then decreased to 0.00005 in the remaining 125\,000 steps using cosine decay~\cite{loshchilovSGDRStochasticGradient2017}. As with the original StyleGAN3, non-saturating logistic loss~\cite{goodfellowGenerativeAdversarialNetworks2020} was used. $R_1$ regularization with $\gamma = 10.0$ was applied to the discriminator after every four training steps~\cite{karrasAnalyzingImprovingImage2020}.

In addition to the generator and discriminator losses, a novel metric called the Frechet U-net distance (FUD) was used to monitor the training process. The FUD is defined as the Fréchet distance of the flattened ASPP outputs of the automatic scoring model~\cite{huttunenComparisonSignalCombinations2023a} between the real and synthetic signals. During training, the FUD was calculated after every 1\,250 training steps using 20\,000 real and 5\,000 synthetic randomly sampled 5-minute segments. The model weights from the training step with the lowest FUD were used for the final model. Similarly to the original StyleGAN3, exponential moving average (EMA) of the generator parameters was tracked during training, with a decay of 0.999~\cite{karrasAliasFreeGenerativeAdversarial2021}. These EMA parameters were then used to create the synthetic dataset.

All deep learning models were implemented with the Jax~\cite{bradburyJAXComposableTransformations2024} and Flax~\cite{heekFlaxNeuralNetwork2024} Python libraries. The models were trained using a single NVIDIA A100 GPU with 40~GB of memory, on a cluster provided by the UEF Bioinformatics Center, University of Eastern Finland. 

\subsection{Assessment of the synthetic data}

To assess the quality of the synthetic signals, a dataset matching the size of the STAGES dataset was generated using the PSG-GAN. The sleep stages and respiratory events were then automatically scored for further analyses. The respiratory events were scored by first finding all consecutive 1-s segments of either hypopnea or apnea. Isolated 1-s segments scored as no event within a longer segment scored otherwise as an event were considered to belong to the event. Finally, the whole event was scored as an apnea if there was at least 10 seconds of apnea present, otherwise it was scored as a hypopnea. Both the synthetic data and STAGES data were scored with the same automatic model to ensure the consistency of the annotations.

The frequency contents of the EEG and EOG signals were analyzed using the relative band powers (delta, theta, alpha, and beta) calculated for each 30-s epoch. First, the power spectral density (PSD) of each epoch was computed with the Welch's method provided by the SciPy~\cite{virtanenSciPy10Fundamental2020} Python package, using 7.5-second Hann window and 50\% overlap. The relative powers for each band were then computed from the PSDs by integrating the frequency bins on each band with the Simpson's rule, and dividing the band powers with the total power. The relative band power distributions were visually compared separately for each sleep stage using violin plots (Fig.~\ref{fig:EEG_EOG_bandpowers}).

To assess the differences in pulse rate variability (PRV) in NREM and REM sleep, the relative LF and HF band powers were calculated from the inter-beat intervals (IBIs) extracted from the PPG peaks in 5-minute windows. The inclusion criteria of a window was that at least 80\% of the window was scored as the same category (wakefulness/NREM/REM). The PPG peaks were automatically detected using the multi-scale peak and trough detection algorithm implemented in the Systole Python package~\cite{legrandSystolePythonPackage2022}. The peaks were detected with 20-second window length and 25\% overlap. After detection, peak artifacts were corrected with Systole's \textit{correct\_peaks} function. The relative LF and HF band powers, and the LF/HF ratio, were calculated with the Systole's \textit{frequency\_domain} function. The relative band power distributions were compared using violin plots (Fig.~\ref{fig:prv_bandpowers}), similarly to the EEG and EOG band powers.

In addition to the PRV band power analyses, the cardiorespiratory interactions were assessed similar to Hietakoste et. al.~\cite{hietakosteLongerApneasHypopneas2020}. The median IBIs were plotted during and after the respiratory events (Fig.~\ref{fig:apnea_hypopnea_ihrs}). Separate plots were produced for apneas vs. hypopneas, and for different event lengths (10--20 s, 20--30 s, and $>$30 s). During each event, 100 points were evenly sampled from the IBI time series with cubic interpolation. After the event, a 15-s segment was sampled from the IBI time series at 4~Hz with cubic interpolation, amounting to 60 sample points. Only those events that had at least 15 seconds between the end of the event and the start of next event were included.

\subsection{Statistical analyses}

The performance of the automatic scoring models was assessed with confusion matrices, overall accuracy, Cohen's kappa ($\kappa$), and the weighted average of class-wise F1-scores. All classification performance metrics were calculated with the Scikit-learn Python package~\cite{pedregosaScikitlearnMachineLearning2011a}.

The distributions of EEG band powers between real and synthetic epochs were visually compared by creating the violin plots with the Seaborn Python package~\cite{waskomSeabornStatisticalData2021}. Median, 25th percentile (Q1), and 75th percentile (Q3) were plotted on top of the violins. Values below $Q1 - 1.5 * IQR$ (inter-quartile range) and above $Q3 + 1.5 * IQR$ were omitted from the violin plots as outliers.

Due to the high number of samples, all differences between real and synthetic data were statistically significant when tested with Kolmogorov-Smirnov and Epps-Singleton tests. For this reason, statistical significance is not reported in the results.



\backmatter





\section*{Acknowledgements}

This research has been conducted using the STAGES - Stanford Technology, Analytics and Genomics in Sleep Resource funded by the Klarman Family Foundation. The investigators of the STAGES study contributed to the design and implementation of the STAGES cohort and/or provided data and/or collected biospecimens, but did not necessarily participate in the analysis or writing of this report. The full list of STAGES investigators can be found at the project website.

The Multi-Ethnic Study of Atherosclerosis (MESA) Sleep Ancillary study was funded by NIH-NHLBI Association of Sleep Disorders with Cardiovascular Health Across Ethnic Groups (RO1 HL098433). MESA is supported by NHLBI funded contracts HHSN268201500003I, N01-HC-95159, N01-HC-95160, N01-HC-95161, N01-HC-95162, N01-HC-95163, N01-HC-95164, N01-HC-95165, N01-HC-95166, N01-HC-95167, N01-HC-95168 and N01-HC-95169 from the National Heart, Lung, and Blood Institute, and by cooperative agreements UL1-TR-000040, UL1-TR-001079, and UL1-TR-001420 funded by NCATS.

The National Sleep Research Resource was supported by the U.S. National Institutes of Health, National Heart Lung and Blood Institute (R24 HL114473, 75N92019R002).

The computational analyses were performed on servers provided by UEF Bioinformatics Center, University of Eastern Finland, Finland.

\section*{Author contribution}

R.H. developed the concept, implemented the models, designed and performed the analyses, and wrote the manuscript. S.K. designed the analyses, supervised the study, and edited the manuscript. M.R. designed the analyses and edited the manuscript. H.K. developed the concept, supervised the study, and edited the manuscript. S.M., J.T., and T.L. supervised the study and edited the manuscript.

\section*{Funding}

This study was funded by the VTR (project numbers 5041789, 5041803, 5041804, 5041820, and 5041822), the Jane and Aatos Erkko Foundation, the Research Council of Finland (project numbers 360119 and 361199), the Finnish Cultural Foundation - Central Fund, the Research Foundation of the Pulmonary Diseases, the Foundation of the Finnish Anti-Tuberculosis Association, and the Emil Aaltonen Foundation.

\section*{Conflict of interest}

The authors declare no conflicts of interest.

\section*{Ethics declarations}

The STAGES dataset is multisite and each participating site has obtained institutional ethics approval for collecting and sharing their data for research use through the National Sleep Research Resource. The collection and use of the MESA dataset was approved by the institutional review boards at each site of the study. The collection and use of the STAR-SA dataset was approved by the Regional Medical Research Ethics Committee of Eastern Finland Collaborative Area (17/2023). The collection and use of the STAR-HV dataset was approved by the Regional Medical Research Committee of the Wellbeing Services County of North Savo (66/2023 and 135/2023).

\section*{Data availability}

The STAGES and MESA datasets are available through the National Sleep Research Resource website. The STAR-SA and STAR-HV datasets are available through SmartSleep laboratory, University of Eastern Finland (Kuopio, Finland) upon reasonable request.

\section*{Code availability}

The code is open source and available at \url{https://github.com/rikuhuttunen/psg-gan}.

\section*{Supplementary information}

Supplementary Video 1: latent space interpolations between randomly sampled latent vectors (huttunen-supplementary-video-1.mp4).










\end{document}